\documentclass[nohyperref]{article}

\usepackage{microtype}
\usepackage{graphicx}
\usepackage{subfigure}
\usepackage{booktabs} 

\usepackage{hyperref}

\usepackage[accepted]{icml2022}

\usepackage{amsmath}
\usepackage{amssymb}
\usepackage{mathtools}
\usepackage{amsthm}

\usepackage[capitalize,noabbrev]{cleveref}

\theoremstyle{plain}

\theoremstyle{definition}

\theoremstyle{remark}

\usepackage[textsize=tiny]{todonotes}

\usepackage{amsmath,amsfonts,bm}
\usepackage{multirow}
\usepackage{multicol}
\usepackage{wrapfig}

\def\gA{{\mathcal{A}}}
\def\gB{{\mathcal{B}}}

\def\gD{{\mathcal{D}}}

\def\gL{{\mathcal{L}}}
\def\gM{{\mathcal{M}}}

\def\gP{{\mathcal{P}}}

\def\gS{{\mathcal{S}}}
\def\gT{{\mathcal{T}}}

\usepackage{xcolor}
\definecolor{MyDarkBlue}{rgb}{0,0.08,1}
\definecolor{MyDarkGreen}{rgb}{0.02,0.6,0.02}
\definecolor{MyDarkRed}{rgb}{0.8,0.02,0.02}
\definecolor{MyDarkOrange}{rgb}{0.40,0.2,0.02}
\definecolor{MyPurple}{RGB}{111,0,255}
\definecolor{MyRed}{rgb}{1.0,0.0,0.0}
\definecolor{MyGold}{rgb}{0.75,0.6,0.12}
\definecolor{MyDarkgray}{rgb}{0.66, 0.66, 0.66}

\newcommand{\yk}[1]{\textcolor{MyDarkRed}{}}
\newcommand{\gc}[1]{\textcolor{MyDarkGreen}{}} 
\newcommand{\sz}[1]{\textcolor{MyDarkBlue}{}} 

\icmltitlerunning{Prompting Decision Transformer for Few-Shot Policy Generalization}

\begin{document}

\twocolumn[
\icmltitle{Prompting Decision Transformer for Few-Shot Policy Generalization}




\begin{icmlauthorlist}
\icmlauthor{Mengdi Xu}{cmu}
\icmlauthor{Yikang Shen}{mila}
\icmlauthor{Shun Zhang}{ibm}
\icmlauthor{Yuchen Lu}{mila}
\icmlauthor{Ding Zhao}{cmu}
\icmlauthor{Joshua B. Tenenbaum}{mit}
\icmlauthor{Chuang Gan}{ibm,umass}
\end{icmlauthorlist}

\icmlaffiliation{cmu}{Carnegie Mellon University}
\icmlaffiliation{ibm}{MIT-IBM Watson AI Lab}
\icmlaffiliation{mit}{Massachusetts Institute of Technology}
\icmlaffiliation{mila}{University of Montreal, Mila}
\icmlaffiliation{umass}{UMass Amherst}

\icmlcorrespondingauthor{Mengdi Xu}{mengdixu@andrew.cmu.edu}

\icmlkeywords{Offline Reinforcement Learning, Decision Transformer, Prompt-based Learning, Few-Shot Learning}

\vskip 0.3in
]



\printAffiliationsAndNotice{}  

\begin{abstract}
Human can leverage prior experience and learn novel tasks from a handful of demonstrations. In contrast to offline meta-reinforcement learning, which aims to achieve quick adaptation through better algorithm design, we investigate the effect of architecture inductive bias on the few-shot learning capability. We propose a Prompt-based Decision Transformer (Prompt-DT), which leverages the sequential modeling ability of the Transformer architecture and the prompt framework to achieve few-shot adaptation in offline RL. We design the trajectory prompt, which contains segments of the few-shot demonstrations, and encodes task-specific information to guide policy generation. Our experiments in five MuJoCo control benchmarks show that Prompt-DT is a strong few-shot learner without any extra finetuning on unseen target tasks. Prompt-DT outperforms its variants and strong meta offline RL baselines by a large margin with a trajectory prompt containing only a few timesteps. Prompt-DT is also robust to prompt length changes and can generalize to out-of-distribution (OOD) environments. Project page: \href{https://mxu34.github.io/PromptDT/}{https://mxu34.github.io/PromptDT/}.
\end{abstract}

\section{Introduction}
\label{sec:intro}

Offline Reinforcement Learning (offline RL)~\citep{levine2020offline} aims to learn an optimal policy from trajectories collected by a set of behavior policies without access to the environments.
This data-driven approach is essential in many settings, where online interactions could be expensive (e.g., robotics or educational agents) and dangerous (e.g., autonomous driving or healthcare). 
A number of recent works have illustrated the power of such approaches in enabling data-driven learning of policies for game environments~\citep{chen2021decision}, robotic manipulation behaviors~\citep{ebert2018visual, kalashnikov2018scalable}, and robotic navigation skills~\citep{kahn2021badgr}.

However, we identify one of offline RL's inherent difficulties as the failure to generalize to unseen tasks.
While the agent might be able to get good state coverage from the training tasks, due to the distribution shift, it would still struggle to find a good policy in the test tasks. 
As a result, recent work from~\citet{mitchell2021offline} considers the offline meta-RL setting that aims to solve the generalization problem in offline RL.
It proposed the Meta-Actor Critic with Advantage Weighting (MACAW) algorithm that uses advantage-weighted regression \cite{peng2019advantage} as a subroutine RL algorithm, and optimizes the agent's adaptive ability with a meta-learning objective \cite{finn2017model}. 



While meta-learning methods address this issue through the algorithmic learning perspective, we aim to investigate the power of architecture inductive bias in this work.
It is known that Transformer~\cite{vaswani2017attention} models, when pretrained on large-scale datasets, are able to perform few-shot or zero-shot learning.
Furthermore, recent works from Natural Language Processing (NLP)~\cite{liu2021pre, brown2020language} suggest the prompt-based framework as an effective paradigm for adaptation to new tasks, 
such as translation and question-answering.
In the prompt-based framework, the prompt contains valuable information about the task, and is pre-pended as a prefix to the input. 
As a result, it casts the problem of few-shot or zero-shot generalization to a conditional sequence generation, which has been the strength of these large Transformer models.
Recently~\citet{chen2021decision} shows that beyond the natural language, the Transformer architecture can also have a strong sequence modeling capability for trajectory data, achieving state-of-art results on offline RL.
In this work, we aim to address the question: \emph{Can we leverage the prompt-based framework from NLP, and adapt it to the context of offline RL to enable few-shot generalization to unseen tasks?}

It is worth noting that adapting the prompt-based method to RL problems is non-trivial.
In NLP, Language Models (LMs) are pretrained on massive amounts of raw text, including almost all information on the internet.
Moreover, most NLP tasks can be rewritten into standard blank-filling formats as prompts.
These prompts serve as queries to extract the right information from the pretrained LMs.
In RL, due to inherent differences between different tasks, it is questionable whether a pretrained model has enough knowledge to solve an unforeseen task.
We propose to use the few-shot demonstrations as prompts, which we call the trajectory prompt.
Instead of unsupervised language model pretraining, we focus on supervisedly training an agent that can imitate these demonstrations to generate a new policy without finetuning.
In our prompt-based offline RL framework, the agent is first trained with offline trajectories that are collected from different tasks in the same domain/environment.
For each task, the agent learns to predict a target trajectory while conditioning on the trajectory prompt sampled from the same task.
During the evaluation, the agent is given a new task and a handful of new trajectories (total step length of which is less or equal to 15) to construct the prompt.
Without extra finetuning, the agent should leverage the task information shown in these trajectories and generate policies for new tasks.
This framework is powerful and attractive for a number of reasons: 
it allows the agent to exploit offline trajectories that are collected from different tasks, 
and the agent can perform few-shot learning, adapting to new scenarios without updating the agent. 

We call our method Prompt-based Decision Transformer (Prompt-DT), which leverages the sequential modeling ability of the Transformer architecture and the prompt framework to achieve few-shot adaptation in offline RL. Our contributions are as follows.
\begin{enumerate}
\item We propose Prompt-DT, a Transformer-based model that learns to adapt to unseen tasks via short trajectory prompts constructed from a handful of trajectories.
\item Our experiments in five MuJoCo control benchmarks show that Prompt-DT is a strong few-shot learner without any extra finetuning on target tasks, beating strong meta offline RL baselines by a large margin. 
\item Our analysis suggests the necessity of our prompt-based framework, as well as the robustness to prompt length and sensitivity to the prompt quality.
\item Prompt-DT can generalize to out-of-distribution tasks, while all the prior methods fail.
\end{enumerate}
\section{Related Work} 
\label{sec:related_works}

\paragraph{Offline Reinforcement Learning.}
Offline RL \cite{levine2020offline} learns a policy with the pre-collected dataset, which contains trajectories sampled under a behavior policy. 
As identified in \citet{levine2020offline}, the offline RL problem has shown to be more challenging than online RL, as the learning agent needs to estimate the value of a policy using only the offline data.
Similar to online RL, we can adopt a model-based or a model-free approach.
When using a model-based approach, we can estimate the reward and transition functions with offline data.
However, we need to modify the RL algorithm to avoid exploiting errors in the estimated model \cite{yu_mopo_2020,kidambi_morel_2021,yu_combo_2021}.
Alternatively, when choosing a model-free approach, we can adapt Q-learning algorithms or the policy gradient algorithms to the offline setting, but need to explicitly correct the distributional mismatch between the behavior policy in the offline data and the policy we want to optimize \cite{kumar_conservative_2020,islam_off-policy_2019}.

\vspace{-0.1in}
\paragraph{Meta-Reinforcement Learning.}
Meta-reinforcement learning (meta-RL) aims to generalize an agent's knowledge from one task to another.
One popular meta-RL algorithm is the Model-Agnostic Meta-Learning (MAML) proposed in \citet{finn2017model}.
The objective of MAML is to find a policy parameter such that given a new task within the task distribution, it can achieve a good performance in the new task only after a few updates.
MAML involves an inner loop and an outer loop in its optimization process.
The inner loop optimizes the policy parameter to adapt to a given task in one step or a few steps. This can be done by following any policy gradient algorithm.
The outer loop involves the meta-learning objective,
which optimizes the performance of the policy {\em after adaption} over all possible tasks in the task distribution.
MAML has shown successful and effective adaptions in benchmark domains.
However, such an optimization algorithm requires backpropagation from the inner loop to the outer loop, which is computationally expensive.
More follow-up methods were proposed to mitigate the computational burden \cite{nichol_first-order_2018,rajeswaran_meta-learning_2019}.



\vspace{-0.1in}
\paragraph{Policy Learning as Sequence Modeling.}
RL algorithms need to handle the challenge of long-term credit assignment, which is typically done by temporal difference (TD) learning \cite{sutton_reinforcement_2018}. 
However, models designed for NLP, like Transformer \cite{vaswani2017attention}, can inherently handle the long-term credit assignment problem.
Recently, Decision Transformer \cite{chen2021decision} was proposed to model an RL problem as a sequence-prediction problem, using state, action, reward-to-go as tokens in a Transformer model.
A concurrent work takes a similar approach that uses Transformer to predict the dynamics of the environment \cite{janner2021reinforcement}. 
These Transformer-based approaches have achieved similar or better performances in benchmark domains compared with classic RL algorithms.
Recently, \citet{furuta2021generalized} demonstrates that Decision Transformer model is doing hindsight information matching.

\vspace{-0.1in}
\paragraph{Few-Shot Learning.}
Few-Shot Learning (FSL) aims to rapidly generalize to new tasks containing only a few samples with supervised information~\citep{wang2020generalizing}.
FSL can advance robotics through developing agents that can replicate human actions. Examples include one-shot imitation~\citep{wu2010towards}, multi-armed bandits~\citep{duan2017one}, visual navigation~\citep{finn2017model}, and continuous control~\citep{yoon2018bayesian}.
Applications of FSL include image classification~\citep{vinyals2016matching}, object tracking~\citep{bertinetto2016learning}, visual question answering~\citep{dong2018fast}, language modeling~\citep{vinyals2016matching}, and neural architecture search~\citep{brock2017smash}.
FSL can reduce the data gathering effort for data-intensive applications.
Another classic FSL scenario is where examples with supervised information are hard to acquire due to safety or ethical issues~\citep{altae2017low}.

\vspace{-0.05in}
\paragraph{Prompt-based Learning.}
For NLP, prompt-based learning is based on language models that model the probability of text directly.
Unlike traditional supervised learning, which trains a model to take in an input $x$ and predict an output $y$ as $P(y|x)$, a prompt-based method uses a template to modify the original input $x$ into a textual string prompt $x'$ that has some unfilled blanks, and then uses the language model to probabilistically fill answers $y$ into blanks~\citep{liu2021pre}.
In this way, by selecting appropriate prompts, we can manipulate the model to predict desired outputs, sometimes even without any additional task-specific training~\citep{brown2020language, radford2019language, gao2020making}.
The underlying hypothesis is that pretrained language models have learned adequate knowledge from the pretraining corpus and we just need to find the right way to extract the knowledge. 
However, in the RL setting, we don't have a pretraining corpus that is large and general enough to cover different environments and tasks. 
Thus, in this work, we propose to use prompts in a different way.
Instead of using prompts to extract knowledge from the pretrained model, the RL agent is required to imitate the provided trajectory prompts, such that it can reproduce the policy that generates these trajectories.




\section{Preliminaries}
\label{sec:preliminary}
\subsection{Online and Offline Meta-Reinforcement Learning}
\label{sec:offline_RL}
A reinforcement learning (RL) problem is a sequential decision-making problem where a learning agent interacts with an environment and optimizes its control policy to obtain the optimal value.
Each sequential decision-making task in dynamic environments is in general modeled as a Markov Decision Process (MDP) \cite{sutton_reinforcement_2018} represented by a tuple $\gM = (\gS, \gA, P, R, \mu)$. $\gS$ and $\gA$ are the state space and the action space. $P: \gS \times \gA \times \gS \rightarrow \mathbb{R}$ is the transition model, where $P(s, a, s')$ is the probability of reaching state $s'$ by taking action $a$ in state $s$. $R: \gS \rightarrow \mathbb{R}$ is the reward function. $\mu$ is the initial state distribution. 
At each step, an RL agent interacts with the environment by taking an action $a$ based on the current state $s$, observing reward $r$ and resulting next state $s'$ from the environment. 
The objective of a sequential decision-making task is to find a policy $\pi: \gS \times \gA \rightarrow \mathbb{R}$ that optimizes the expected cumulative rewards, $\mathbb{E}_{s_0 \sim \mu, \pi} \sum_t\gamma^t R(s_t)$.

Generally, RL is performed online, where the agent iteratively takes actions and receives feedback from the environment.
However, this may not always be feasible as RL algorithms may require a large number of training data due to their generally low sample efficiency.
This makes training in an online environment time-consuming.
In some real-world safety-critical environments, deploying the agent online in the training phase may cause catastrophic failures.
We consider the {\em offline RL} setting \cite{levine2020offline}, which aims to learn a policy from data that are pre-collected using a (possibly-unknown) behavior policy.
In this setting, the agent has access to a dataset $\gD$ containing a set of trajectories. A trajectory $\{s_0, a_0, r_0, s_1, a_1, r_1, \dots, s_T, a_T, r_T\}$ is sampled using a behavior policy in the environment. 
The agent is expected to find the optimal policy using only the dataset $\gD$ without interacting with the environment itself.

Both online and offline RL settings are originally proposed to find the optimal policy in one task.
The efficiency of RL can be further improved if the designed learning agent is able to adapt to similar tasks with a handful of newly collected data after learning on a few tasks, which is mainly developed under {\em meta-RL} \cite{finn2017model}.
Recently, {\em meta-RL} has been extended to offline settings, aiming to adapt to new tasks via pre-collected data quickly.
In the {\em offline meta-RL} setting proposed in \citet{mitchell2021offline}, an agent is given a set of tasks $\gT$, where a task $\gT_i \in \gT$ is defined as $(\gM_i, \pi_i)$, containing an MDP $\gM_i$ and a behavior policy $\pi_i$.
For each task $\gT_i$, the agent is provided with a dataset $\gD_i$, which contains trajectories sampled using $\pi_i$.
The agent is trained with a subset of training tasks denoted as $\gT^{train}$ and is expected to find the optimal policies in a set of test tasks $\gT^{test}$, which is disjoint with $\gT^{train}$.

\begin{figure*}[tb]
    \begin{center}
        \centerline{\includegraphics[width=\linewidth]{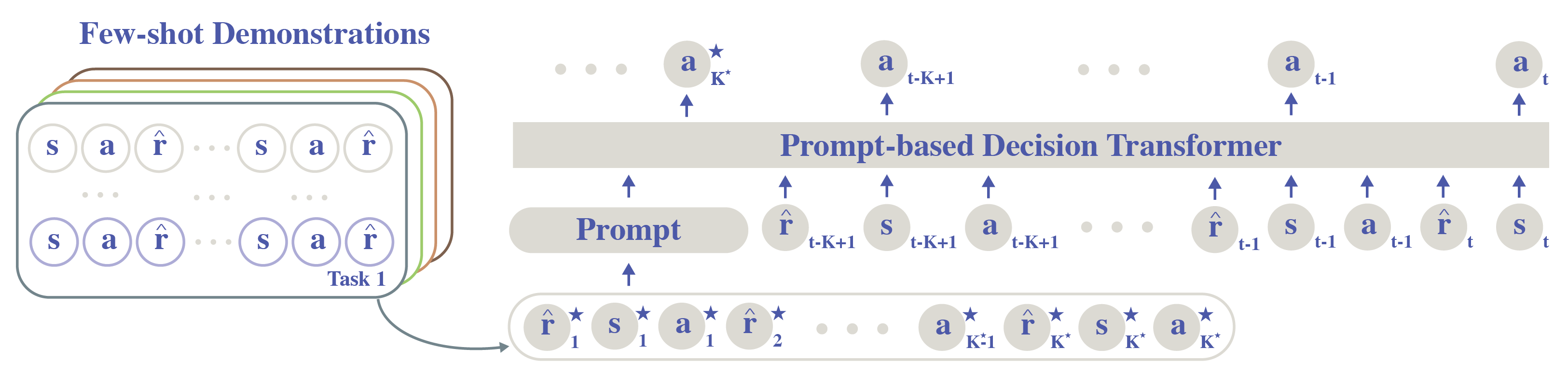}}
        \vskip -0.15in
        \caption{Prompt-DT for few-shot policy generalization. The left shows the few-shot demonstration dataset $\gP_i$ for each task $\gT_i \in \gT^{train} \cup \gT^{test}$. The trajectory prompt is defined as a trajectory sequence of length $K^{\star}$ sampled from various episodes stored in $\gP_i$. In both pretraining and few-shot evaluation, Prompt-DT takes both the trajectory prompt augmentation and the most recent $K$-step history as input, and autoregressively outputs actions corresponding to each state in the input sequence.}
        \label{fig:prompt_dt}
        \end{center}
        \vspace{-0.2in}
\end{figure*}

\subsection{Decision Transformer}
\label{sec:dt}
Transformer which has been extensively studied in NLP \cite{devlin2018bert} and computer vision \cite{carion2020end}, has shown to outperform RNN-based architectures.
It is recently applied to solve RL problems for its efficiency and scalability when modeling long sequential data.
Decision Transformer \cite{chen2021decision} for offline RL treats learning a policy as a sequential modeling problem. 
It proposes to model trajectories with state $s_t$, action $a_t$ and reward-to-go $\hat{r}_t$ tuples collected at different time steps $t$.
The reward-to-go is the cumulative rewards from the current time step till the end of the episode. 
Instead of including the one-step reward $r_t$, this novel representation helps guide action selection towards optimizing the return.
At timestep $t$, Decision Transformer takes a trajectory sequence $\tau$ autoregressively as input which contains the most recent $K$-step history.
\begin{align}
    \tau = (\hat{r}_{t-K+1}, s_{t-K+1},a_{t-K+1}, 
    \dots, \hat{r}_{t},s_{t}, a_{t}).
\end{align}

When training with offline collected data, $\hat{r}_t = \sum_{i=t}^T r_i$. During testing, $\hat{r}_t = G^{\star} - \sum_{i=0}^t r_i$ where $G^{\star}$ is the targeted total return for an episode.
Each trajectory $\tau$ corresponds to $3K$ tokens in the standard Transformer model.
To encode the sequence timestep information, Decision Transformer concatenate the same timestep embedding to the embeddings of $s_t$, $a_t$ and $\hat{r}_t$.
Each head corresponding to a state token is trained to predict an action by minimizing mean-squared loss when continuous action spaces.



\section{Prompt-based Decision Transformer}
\label{sec:method}

This section presents Prompt-based Decision Transformer (Prompt-DT), a novel Transformer architecture for few-shot policy generalization as visualized in Figure~\ref{fig:prompt_dt}. 

\subsection{Problem Formulation} 
We formalize the offline few-shot RL problem as a few-shot policy generalization problem to new tasks after training on a set of tasks with offline data. 
Each task $\gT_i$ in the training set $\gT^{train}$ is associated with a dataset $\gD_i$, which contains trajectories pre-collected with an unknown behavior policy $\pi_i$. 
In contrast to offline meta-RL, which updates the model weights with task-specific offline data or online interactions, we desire to achieve generalization with no finetuning or gradient updates, which maintains high efficiency and avoids catastrophic forgetting due to parameter shifts. 

To achieve few-shot learning in the context of RL, we assume that there exists a dataset $\gP_i$ containing few-shot demonstrations for each task $\gT_i \in \gT^{train} \cup \gT^{test}$.
For a training task $\gT_i  \in \gT^{train}$, we let $\gP_i$ be a subset of the offline data set $\gD_i$. 
$\gP_i$ for a test task $\gT_i  \in \gT^{test}$ could be obtained with a human experimenter or a behavior policy.
In this work, we hope to design an architecture that can directly extract unique task-specific information stored in the demonstration dataset $\gP_i$ and exploit the information to guide policy generation.


\subsection{Prompt Representation} 
Text prompts containing task-specific instructions enable a large Transformer model to generate answers without changing the model parameters in NLP tasks.
In the context of RL, text descriptions that could serve as prompts are recently introduced to solve Atari video games and multi-modal household tasks \cite{shridhar2020alfred}.  
Such text descriptions usually require predefined language templates and may require large human labor to annotate.

We instead define {\em trajectory prompt} for RL as a sequence that consists of a few trajectory segments.
Each trajectory segment contains multiple state $s^{\star}$, action $a^{\star}$ and reward-to-go $\hat{r}^{\star}$ tuples, ($s^{\star}, a^{\star}, \hat{r}^{\star}$), following the trajectory representation in Decision Transformer. 
Each element with superscript ${\cdot}^{\star}$ is associated with the trajectory prompt.
Since each sequential decision-making task $\gT_i$ can be modeled as an MDP $\gM_i$, trajectory prompts can store partial to complete information to specify a task by implicitly capturing the transition model and the reward function. 
Trajectory prompts are relatively easy to obtain compared to text prompts by directly sampling trajectory segments from the few-shot demonstration dataset $\gP_i$.
Formally, we define a trajectory prompt $\tau^{\star}_i$ for task $\gT_i$ as 
\begin{align}
    \tau^{\star}_i = (\hat{r}^{\star}_{1}, s^{\star}_{1},a^{\star}_{1},
    \hat{r}^{\star}_{2}, s^{\star}_{2},a^{\star}_{2}, 
    \dots, \hat{r}^{\star}_{K^{\star}},s^{\star}_{K^{\star}},a^{\star}_{K^{\star}} ).
\end{align}
$K^{\star}$ is the number of environment steps stored in the prompt.
Note that our choice of $K^{\star}$ is much shorter than the horizon of the task.
So a trajectory prompt only contains the information needed to help identify the task but insufficient information for the agent to imitate.

\subsection{Prompt-DT Architecture}
Our Prompt-DT architecture is built on Decision Transformer \cite{chen2021decision} and solves the offline few-shot RL problem through the lens of a prompt-augmented sequence-modeling problem.  
The proposed trajectory prompt allows minimal architecture change to the Decision Transformer for generalization.
For each task $\gT_i$, Prompt-DT takes $\tau^{input} $ as input, which contains both the $K^{\star}$ step trajectory prompt obtained from $\gP_i$ and the most recent $K$ step history sampled from $\gD_i$. Formally $ \tau^{input} = (\tau_i^{\star}, \tau_i)$.
Since the data pair at each timestep is a 3-tuple $(s,a, \hat{r})$, the input sequence corresponds to $3(K^{\star}+K)$ tokens in Transformer. 
Prompt-DT autoregressively outputs $K^{\star}+K$ actions at heads corresponding to state tokens in the input sequence.


In the implementation, we utilize stochastic trajectory prompt $\tau^{\star}_i$ aiming to increase training stability and avoid overfitting as in Algorithm~\ref{alg:getprompt}.
The stochastic trajectory prompt $\tau^{\star}_i$ for task $\gT_i$ consists of $J$ trajectory segments with length $H$ and $K^{\star} =JH$. 
\begin{align}
    \tau^{\star}_i = & \ (\tau^{\star}_{i,1}, \dots, \tau^{\star}_{i,J} ), \  \\
    \tau^{\star}_{i,j} = & \ (\hat{r}^{\star}_{i,j,1}, s^{\star}_{i,j,1},a^{\star}_{i,j,1}, \dots, \nonumber \\
    & \quad \quad  \hat{r}^{\star}_{i,j,H},s^{\star}_{i,j,H},a^{\star}_{i,j,H}), \forall j \in [J]. \label{eq:segment}
\end{align}
Following the structure of Decision Transformer, we utilize a GPT model with linear layers to obtain token embeddings and add the same positional embedding to tokens corresponding to the same timestep.



\begin{algorithm}[tb]
    \caption{Prompt-DT Training}
    \label{alg:training}
\begin{algorithmic}
    \STATE {\bfseries Input:} training tasks $\gT^{train}$, causal Transformer \textit{Transformer}$_{\theta}$, training iterations $N$, offline dataset $\gD$, demonstrations $\gP$, per-task batch size $M$, learning rate $\alpha$
    \FOR{$n=1$ {\bfseries to} $N$}
    \FOR{Each task $\gT_i \in \gT^{train}$}
    \FOR{$m=1$ {\bfseries to} $M$}
    \STATE Sample a trajectory $\tau_{i,m}$ of length $K$ from $\gD_i$
    \STATE Sample a prompt $\tau^{\star}_{i,m} =$ \textit{GetPrompt}($\gT_i, \gP_i$)
    \STATE Get input $\tau^{input}_{i,m} = (\tau_{i,m}^{\star}, \tau_{i,m})$
    \ENDFOR
    \STATE Get a minibatch $ \gB_{i}^M = \{\tau^{input}_{i,m} \}_{m=1}^M$
    \ENDFOR
    \STATE Get a batch $ \gB = \{ \gB_{i}^M \}_{i=1}^{T^{train}}$, $T^{train} = |\gT^{train}| $
    \STATE $a^{pred}$ = \textit{Transformer}$_{\theta}(\tau^{input})$, $\forall \tau^{input} \in \gB$
    \STATE $\gL_{MSE} = \frac{1}{|\gB|} \sum_{\tau^{input} \in \gB}(a - a^{pred})^2$
    \STATE $\theta  \leftarrow \theta - \alpha \nabla_{\theta} \gL_{MSE}$
    \STATE Prompt-DT Few-Shot Evaluation along training
    \ENDFOR
\end{algorithmic}
\vspace{-0.01in}
\end{algorithm}

\subsection{Algorithms}
We summarize the training and testing algorithms for Prompt-DT in Algorithm~\ref{alg:training} and Algorithm~\ref{alg:testing}. 

\vspace{-0.05in}
During training, Prompt-DT minimizes errors between the predicted actions and the actions in the data for both the prompt and recent history.
By learning the target actions stored in trajectory prompts, Prompt-DT is motivated to extract the task-specific information stored in the trajectory prompt and combine it with the recent history for future action predictions. 
In continuous settings, Prompt-DT minimizes the mean-squared loss with gradient descent.
At each training step, we sample a batch $\gB$ that contains prompt-history pairs for \textit{each} training task.
Instead of iteratively conducting gradient updates with data batch sampled from one training task, our batch $\gB$ helps stabilize training by aggregating gradient estimates across all tasks in $\gT^{train}$.

In testing time, we assume that the offline pretrained Prompt-DT can interact with a simulator for each evaluation task in $\gT^{test}$. 
This online evaluation assumption resembles the actual deployment of trained RL agents. 
We sample a stochastic trajectory prompt based on the demonstration set $\gP_i$ for task $\gT_i \in \gT^{test}$ similar to the training procedure. Prompt-DT then generates actions taking both the prompt and recent context as input.
At the beginning of each episode, we initialize a recent history $\tau$ with all zeros for Prompt-DT prediction and update it with streamingly collected data.

\setlength{\textfloatsep}{0pt}
\begin{algorithm}[tb]
    \caption{Trajectory Prompt Generation (\textit{GetPrompt})}
    \label{alg:getprompt}
\begin{algorithmic}
    \STATE {\bfseries Input:} task $\gT$, task-specific demonstrations $\gP$, sample episode $J$, segmentation length $H$
    \STATE Sample $J$ episodes from $\gP$
    \STATE Sample segments $\tau^{\star}_j$ of length $H$, $\forall j \in [J]$ (Equation~\ref{eq:segment})
    \STATE {\bfseries Return:} trajectory prompt $\tau^{\star} = (\tau^{\star}_{1}, \dots, \tau^{\star}_{J} )$
    
\end{algorithmic}

\end{algorithm}
\setlength{\textfloatsep}{1cm}
\begin{algorithm}[tb]
    \caption{Prompt-DT Few-Shot Evaluation}
    \label{alg:testing}
\begin{algorithmic}
    \STATE {\bfseries Input:} test tasks $\gT^{test}$, \textit{Transformer}$_{\theta}$, demonstrations $\gP$, target return $G^{\star}$, episode length $T$
    \FOR{Each task $\gT_i \in \gT^{test}$}
    \STATE Initialize history $\tau$ with zeros, desired reward $g = G^{\star}_i$
    \STATE Sample a prompt $\tau^{\star} =$ \textit{GetPrompt}($\gT_i, \gP_i$)
    \FOR{$t\leq T$}
    \STATE Get action $a$ = \textit{Transformer}$_{\theta}( (\tau^{\star}, \tau) )$[-1]
    \STATE Step env. and get feedback $s, a, r$, $g \leftarrow g-r$
    \STATE Append $[s, a, g]$ to recent history $\tau$
    \ENDFOR
    \ENDFOR
\end{algorithmic}
\vspace{-0.01in}
\end{algorithm}

\begin{figure*}[t]
    \begin{center}
    \centerline{\includegraphics[width=\linewidth]{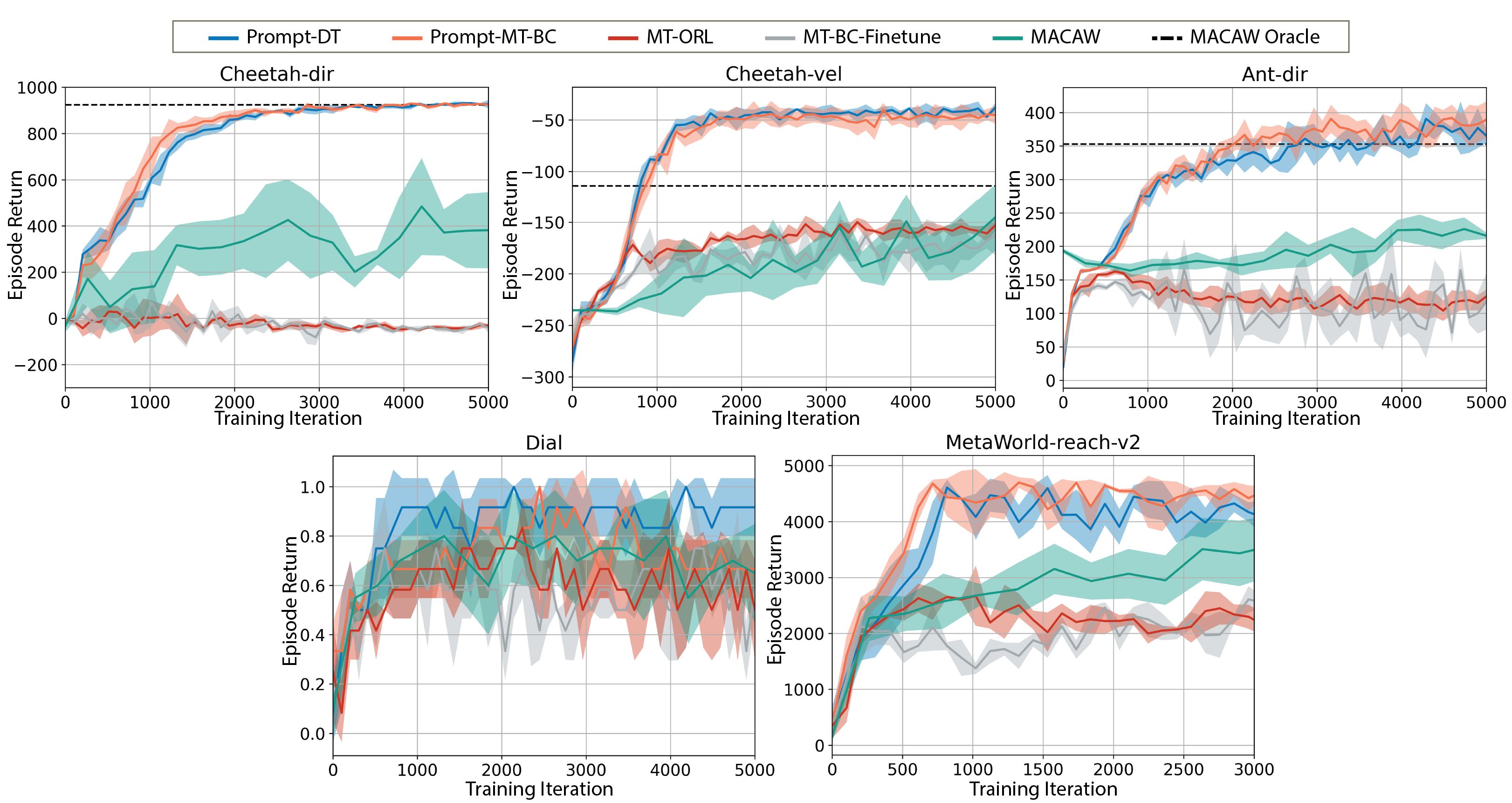}}
    \vspace{-0.15in}
    \caption{Episodic accumulated returns in never-before-seen tasks of Prompt-DT, Prompt-based Behavior Cloning (Prompt-MT-BC), Multi-task Offline RL (MT-ORL), Multi-task Behavior Cloning (MT-BC-Finetune), and Meta-Actor Critic with Advantage Weighting (MACAW). All methods are trained with the same expert dataset $\gD$. Each plot is run with three seeds. Prompt-DT and Prompt-MT-BC have a few-shot dataset $\gP$ containing expert demonstrations. Cheetah-dir, Cheetah-vel and Ant-dir have prompts of length $K^{\star}=5$. Dial has prompts of length $K^{\star}=15$. Meta-World reach-v2 has prompts of length $K^{\star}=2$. MT-BC-Finetune and MACAW use the same amount of data which equals the prompt length for finetuning at testing time.
    The dashed lines show the optimal performance of MACAW reported in \citet{mitchell2021offline}.
    Prompt-augmented methods including Prompt-DT and Prompt-MT-BC outperform baselines across environments with a short trajectory prompt.
    }
    \label{fig:main_comparison_expert}
    \end{center}
    \vspace{-0.2in}
\end{figure*}

\vspace{-0.05in}
\section{Experiments}
\label{sec:experiment}
\vspace{-0.05in}
We conduct experiments to test the few-shot generalization capability of the proposed Prompt-DT with metric as the episode accumulated reward. 
We aim to empirically answer the following questions:
1) Can Prompt-DT achieve few-shot policy generalization?
2) How does the prompt quantity and quality affect the few-shot generalization ability?
3) Does Prompt-DT generalize to out-of-distribution tasks?

\vspace{-0.05in}
\subsection{Environments and Datasets}
\label{sec:env_data}
\vspace{-0.05in}
We evaluate in five meta-RL control tasks described as follows \cite{finn2017model,rothfuss2018promp, mitchell2021offline, yu2020meta, todorov2012MuJoCo}.

\begin{itemize}
\vspace{-0.05in}
\item \textbf{Cheetah-dir.} There are two tasks in Cheetah-dir with goal directions as forward and backward, respectively. The cheetah agent is rewarded with high velocity along the goal direction. The training and testing set are equal, and both contain the two tasks.
\vspace{-0.05in}
\item \textbf{Cheetah-vel.} There are 40 tasks in Cheetah-vel with different goal velocities. The target velocities are uniformly sampled from the interval [0,3]. The agent is penalized with $l_2$ errors to the target velocity. We hold out 5 tasks to construct the testing set and train with the remaining 35 tasks. 
\vspace{-0.05in}
\item \textbf{Ant-dir.} There are 50 tasks in Ant-dir with different goal directions uniformly sampled in 2D space. The 8-joints ant is rewarded with high velocity along the goal direction. We sample 5 tasks for testing and leave the rest for training. 
\vspace{-0.05in}
\item \textbf{Dial.} In Dial, the task is to control a 6-DOF Jaco robot to reach a target number located in a number pad. There are 10 numbers in total corresponding to 10 tasks. We train in 6 tasks and test in 4 tasks. Dial is more complex than Meta-World reach-v2 since it directly controls 6 joints.
\vspace{-0.05in}
\item \textbf{Meta-World reach-v2.} In Meta-World reach-v2, the task is to control a Sawyer robot's end-effector to reach a target position in 3D space. The agent directly controls the XYZ location of the end-effector. Each task has a different goal position. We train in 15 tasks and test in 5 tasks.
\end{itemize}
\vspace{-0.12in}
Experiments in Cheetah-dir, Cheetah-vel and Ant-dir strictly follow the datasets and settings in \citet{mitchell2021offline}. 
The agents in all three tasks are penalized with large control signals.
The dataset for Cheetah-dir and Ant-dir contains the full replay buffer for training an RL agent with Soft Actor-Critic \cite{haarnoja2018soft}. The dataset for Cheetah-vel contains the replay buffer trained with TD3 \cite{fujimoto2018addressing} for its better training stability.
In Meta-World reach-v2 \cite{yu2020meta} and Dial \cite{shiarlis2018taco}, we collect expert trajectories with script expert policies provided in both environments.

\subsection{Baselines}
\label{sec:baselines}
We compare the few-shot generalization ability of \textbf{Prompt-DT} with four baselines, including three variants of Prompt-DT and one state-of-art offline meta-RL algorithm.
\begin{itemize}
\vspace{-0.1in}
\item \textbf{Multi-task Offline RL (MT-ORL).} We train a Decision Transformer to learn multiple tasks in the training set. We omit the prompt augmentation in Prompt-DT to construct MT-ORL and keep the remaining MT-ORL training process the same as Prompt-DT. In evaluation, the reward-to-go are fed into the Transformer model to provide partial task-specific information. MT-ORL helps ablate the effect of prompt.
\item \textbf{Prompt-based Behavior Cloning (Prompt-MT-BC).} We omit Prompt-DT's reward-to-go tokens stored in the history input in both training and evaluation. This Prompt-MT-BC baseline only keeps task-specific information in the trajectory prompt. 
Prompt-MT-BC helps show the effect of reward-to-go tokens.
\item \textbf{Multi-task Behavior Cloning (MT-BC-Finetune).} We exclude both the prompt augmentation and reward-to-go tokens in the MT-BC-Finetune baseline. To adapt to the target task, we update the Decision Transformer model with finetuning gradient steps with data collected in the target task. MT-BC-Finetune helps show the effect of both prompt and reward-to-go tokens compared with finetuning.
\item \textbf{Meta-Actor Critic with Advantage Weighting (MACAW).} MACAW is an offline meta-RL algorithm that leverages the strength of both meta-learning and off-policy value-based algorithms. MACAW has high sample efficiency and outperforms multiple finetune adaptation baselines in Cheetah-dir, Cheetah-vel, and Ant-dir \cite{mitchell2021offline}. 
\end{itemize}




\section{Discussion}
\label{sec:discussion}





\subsection{Can Prompt-DT Achieve Few-Shot Policy Generalization?}
\label{sec:exp_prompt_main}

We compare the few-shot generalization ability of Prompt-DT and the baselines to investigate whether prompts facilitate few-shot generalization, whether prompts encode adequate task-specific information than the rewards, and whether prompt-augmented methods are more data-efficient than finetuning methods. 
We measure the generalization ability of different methods with the average episode accumulated reward in $\gT^{test}$.
We show the results in Figure~\ref{fig:main_comparison_expert}, which are the few-shot evaluation performances of different algorithms along the training process.

All the methods in Figure~\ref{fig:main_comparison_expert} are trained with the same expert dataset in each environment.
For Cheetah-dir, Cheetah-vel, and Ant-dir, we select the expert dataset for each task as the last $20,000$ steps in the complete replay buffer.  
The expert dataset for Dial and Meta-World reach-v2 contains 200 episodes for each task.
Prompt-DT and Prompt-MT-BC make use of a few-shot dataset $\gP$ containing expert demonstrations sampled from $\gD$, and have a trajectory prompt consisting of a segment sampled from a single episode. 
To have a fair comparison, finetuning methods, including MT-BC-Finetune and MACAW, use offline expert trajectories for estimating finetune gradients and are evaluated in an online manner. 
We later discuss the effect of demonstration dataset $\gP$'s quantity in Section~\ref{sec:ablation_prompt_length}, and dataset $\gD$ and demonstration dataset $\gP$'s quality in Section~\ref{sec:ablation_prompt_quality}.

\paragraph{Comparison of MT-ORL and Prompt-MT-BC.}
\label{sec:ablation_importance_prompt}
With expert datasets $\gP$ and $\gD$, Prompt-DT achieves high performance with few-shot demonstrations in unseen tasks as shown in Figure~\ref{fig:main_comparison_expert}.
Prompt-DT consistently outperforms MT-ORL across environments with a large margin. Prompt-DT and Prompt-MT-BC perform similarly in Cheetah-dir, Cheetah-vel, Ant-dir, and Meta-World reach-v2. 
This observation shows that the trajectory prompts already embed sufficient information to fully specify the task. 
However, Prompt-DT outperforms Prompt-MT-BC in Dial,
which shows that there exist environments where the prompt itself is insufficient, and the rewards help in policy generalizations.

Prompt-MT-BC performs similarly to MT-ORL in Dial and is better than MT-ORL in the other four environments by a large margin. 
We can see that the expert trajectory prompt augmentation indeed provides more task-specific information than the reward-to-go tokens stored in the historical context in most situations.
In Cheetah-dir, Cheetah-vel and Ant-dir, although we construct different tasks by only modifying reward functions (e.g., changing goal directions and velocities), it is still hard to directly generalize to novel tasks with contexts containing recent reward records according to the poor performances of MT-ORL. In contrast, the trajectory prompt augmentation could provide strong task-specific signals to guide the action generation.



\paragraph{Comparison of MACAW and MT-BC-Finetune.}
\label{sec:ablation_finetune}
During training, we use the same batch sizes for all the algorithms. Thus, methods at the same number of training iterations (the x-axis in Figure~\ref{fig:main_comparison_expert}) use the same amount of training data. During testing, we provide all methods with the same amount of data collected in the target test task, which means the amount of data used for finetuning equals the prompt length for each environment. 

In all the environments except for Dial, Prompt-based methods converge to the optimal performances much faster than MACAW.
Note that the optimal performance of MACAW marked as the dashed line in Figure~\ref{fig:main_comparison_expert} requires a finetune batch of size 256.
Even with a much smaller amount of adaptation data (5 vs. 256), prompt-based methods have better asymptotic performance than MACAW in relatively complex environments, including Cheetah-vel and Ant-dir, and similar performance in the simple Cheetah-dir environment.
In all the environments, prompt-based methods consistently outperform MT-BC-Finetune. 
Moreover, with the same amount of finetuning data and finetuning steps, MT-BC-Finetune underperforms MACAW.
We conjecture that this is due to the data-hungry nature of Transformer, which has significantly more model parameters than the actor and critic net of MACAW.
We provide further ablation studies to show the performance of finetune-based algorithms with various amounts of finetune data and finetune steps in Section~\ref{sec:ablation_finetune_data_MT_BC_Finetune} and Section~\ref{sec:ablation_finetune_data_MACAW}.

\subsection{Does the Prompt Quantity Affect the Few-Shot Generalization Ability?}
\label{sec:ablation_prompt_length}
In practice, there may exist a limited amount of high-quality demonstrations for each test task, or the demonstrations may contain trajectories with heterogeneous quality. 
We provide an ablation study to reveal the effect of the trajectory prompt quantity on the few-shot generalization ability.
Table~\ref{tab:ablation_prompt_length} summarizes the ablation results with expert dataset $\gD$ and expert demostrations $\gP$ in all the three environments.
We vary the prompt quantity $K^{\star}$ by changing the number of trajectory segments $J$ and the segment length $H$.

\begin{table}[tb]
\caption{Ablation: The effect of prompt quantity on Prompt-DT's few-shot generalization ability. We vary the number of episodes $J$ and the trajectory segment length $H$ in three environments. We use a recent history containing $K=20$ timesteps for all experiments. Each number is run with 3 seeds. Prompt-DT only requires an expert prompt with a small number of timesteps to achieve a few-shot generalization. }
\vspace{0.1in}
\label{tab:ablation_prompt_length}
\centering
\begin{tabular}{cccccc}
  \toprule
  $K^{\star}$ &J & H & Cheetah-dir & Cheetah-vel & Ant-dir \\
  \midrule
  2 & 1  & 2 & 926.46 $\pm$ 2.87 & -45.26 $\pm$ 2.87 & \textbf{409.81 $\pm$ 9.69} \\
  5 & 1  & 5  & \textbf{927.20 $\pm$ 18.02} &  -37.92 $\pm$ 4.56 & 367.12 $\pm$ 10.50\\
  10 & 1  & 10 &  925.00 $\pm$ 1.41 & -38.43 $\pm$ 2.14 & 382.94 $\pm$ 25.21 \\
  40 & 2  & 20 & 926.87 $\pm$ 9.30 &  \textbf{-34.43 $\pm$ 2.33} & 323.83 $\pm $ 9.33 \\
  \bottomrule
\end{tabular}
\vspace{-0.1in}
\end{table}

In Cheetah-dir, Prompt-DT achieves similar high performances with different prompt lengths, even when the prompt only contains two timesteps. It shows that by training in both tasks in Cheetah-dir, Prompt-DT learns to run fast with history context $\tau$, which is a common skill shared across tasks, and extracts goal direction information stores in the $(s, a, \hat{r})$ pair to generate actions. 
In Cheetah-vel, the trajectory prompt with parameters $K^{\star}=40, J=2, H=20$ achieves the highest episode return in unseen task sets. 
Increasing the prompt length $K^{\star}$ does not greatly increase the generalization performance.
With a short two-timestep trajectory prompt, Prompt-DT could still achieve higher performances than other baselines. 
In Ant-dir, Prompt-DT achieves the highest performance with a two-timestep prompt. We conjecture that the existence of nonexpert episodes (with negative returns) in Ant-dir's dataset decreases the effectiveness of prompts constructed from multiple trajectories.

Our experiments show that, with trajectory prompt sampled from expert demonstrations, Prompt-DT is not sensitive to the prompt quantity and can successfully extract task-specific information even with prompts containing only a few timesteps. It shows that the sequential information in the trajectory prompt is not crucial to revealing the unique task-specific reward information in the current MuJoCo control settings. However, we conjecture that if tested with a more complex task set, such as Meta-world containing tasks that require sequential tool manipulation, the sequential information stored in the prompt will become more critical.

\begin{figure*}[tb]
    \begin{center}
    \centerline{\includegraphics[width=\linewidth]{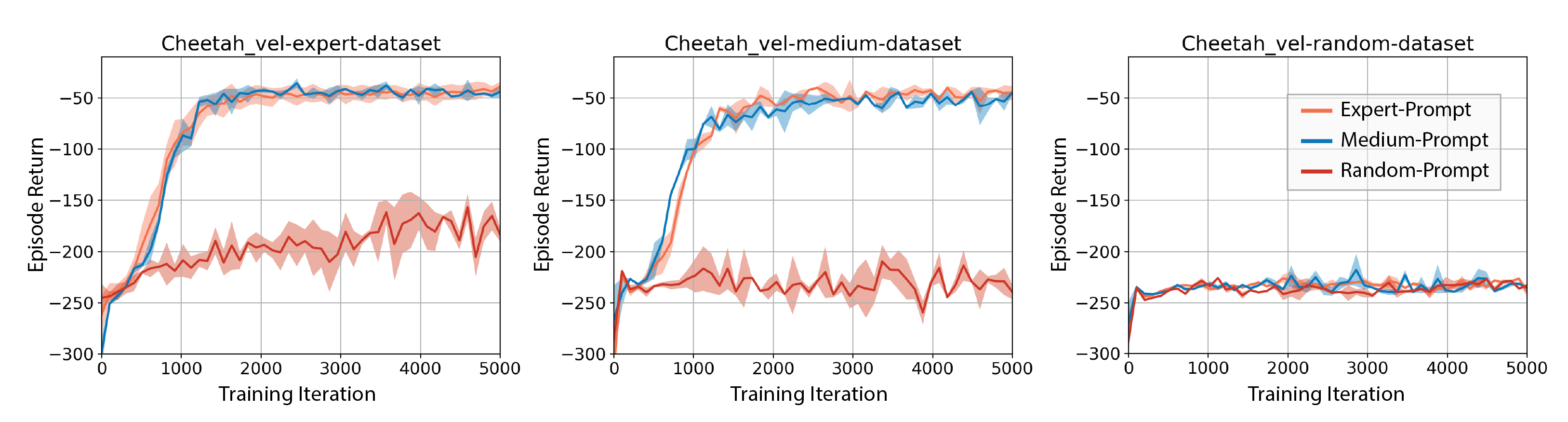}}
    \vspace{-0.15in}
    \caption{Ablation: The effect of prompt quality to Prompt-DT's few-shot generalization ability. We train Prompt-DT with datasets and demonstrations with the same quality in each plot. The left, middle and right figure corresponds to expert, medium, and random dataset collected in Cheetah-vel. Each plot is run with 2 seeds. We feed Prompt-DT trajectory prompts of different qualities when testing Prompt-DT's few-shot generalization ability. The results show that Prompt-DT tries to generate policies that match the prompt quality, and the quality of training datasets also affects the few-shot generalization ability of Prompt-DT. }
    \label{fig:Ablation_cheetah_vel_quality}
    \end{center}
    \vspace{-0.2in}
\end{figure*}

\subsection{Does the Prompt Quality Affect the Few-Shot Generalization Ability?}
\label{sec:ablation_prompt_quality}

There are situations where expert demonstrations for target tasks are unavailable, especially when the target test task $\gT$ itself is not fully characterized due to model uncertainty (e.g., uncertain dynamic parameters in control tasks). 
Thus it is vital to investigate how the prompt quality affects the generalization ability. 
We conduct an ablation study in Cheetah-vel and construct expert, medium, and random datasets $\gD'$ corresponding to the last, middle, and first 500 trajectories in the full replay buffer collected along training an RL agent. We also sample expert, medium, and random few-shot demonstrations $\gP'$ from $\gD'$ accordingly. We summarize the ablation results in Figure~\ref{fig:Ablation_cheetah_vel_quality}.

Prompt-DT could adjust its generated actions according to the given trajectory prompt when training with expert data. For example, Prompt-DT achieves low episode return with random trajectory prompt and high performances with expert or medium prompts.
Similarly, when training with medium data, Prompt-DT's return heavily decreases when random prompts and slightly increases with expert prompts, validating the effect of prompt quality. 
However, when training with random data, only feeding Prompt-DT expert or medium trajectory prompts does not help improve the generalization ability. 
This observation may result from the largely overlapped task state distributions in the random training dataset, reducing the prompt's effects and encouraging Prompt-DT to match the commonly shared random state distribution. 





\begin{figure}[t]
    \begin{center}
    \centerline{\includegraphics[width=0.9\linewidth]{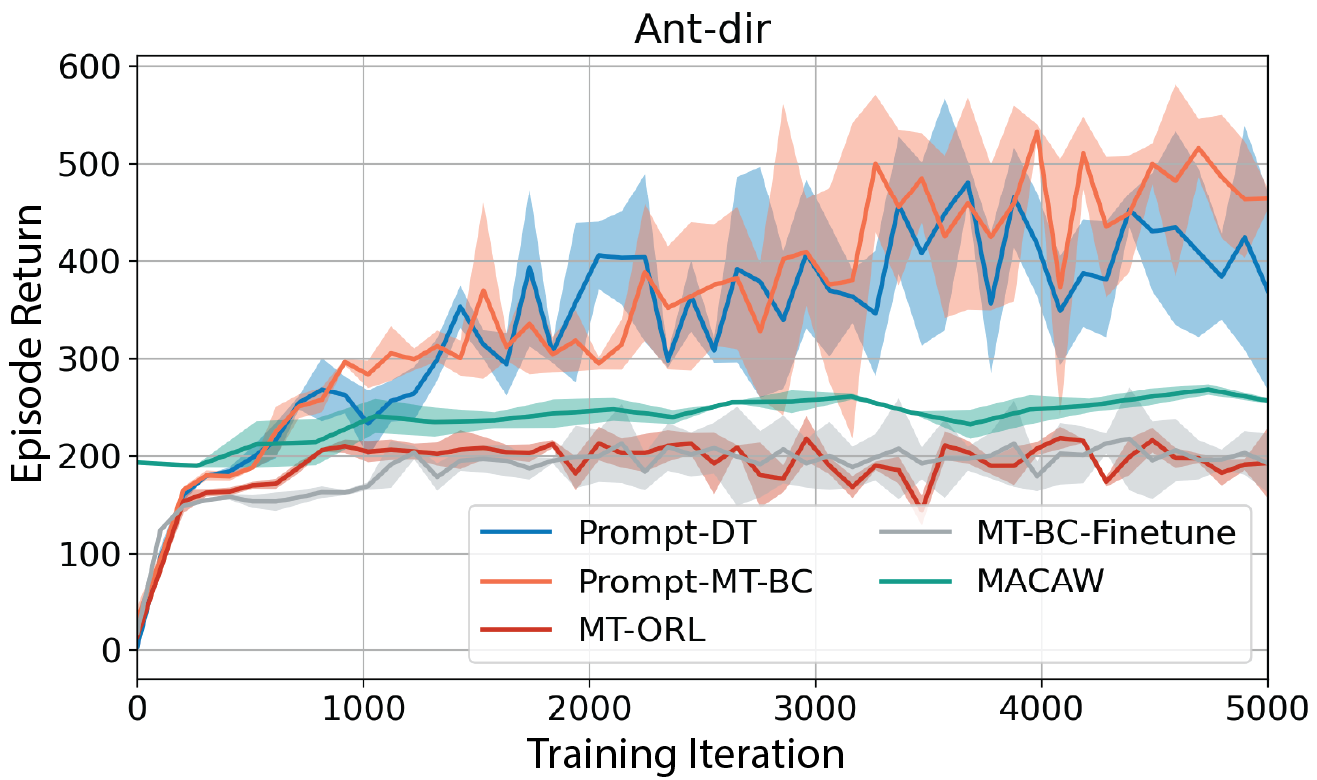}}
    \vspace{-0.15in}
    \caption{Episodic accumulated returns in novel tasks with goals out of training tasks' goal range in Ant-dir. Each plot is run with 3 seeds.
    Prompt-MT-BC scores the highest reward and outperforms baselines without trajectory prompts by a large margin.
    }
    \label{fig:Ablation_ant_dit_ood}
    \end{center}
    \vspace{-0.2in}
\end{figure}

\subsection{Can Prompt-DT Generalize to Out-of-distribution Tasks via Few-Shot Demonstrations?}
\label{sec:ablation_ood}

In previous discussions, we follow the experimental setting in MACAW \cite{mitchell2021offline} by training in a large batch of training tasks and evaluating in a few held-out tasks. The held-out tasks have goals (target velocity or target direction) within the goal range calibrated by training tasks. We desire to test whether trajectory prompts enable the extrapolation ability when handling tasks with goals out of the training range. In other words, we hope to test the generalization ability in out-of-distribution tasks.

We sample eight training tasks in Ant-dir and three testing tasks, two of which have indexes smaller than the minimum task index and one larger than the maximum. The task index is proportional to the desired direction angle. We show that Prompt-DT still performs better than baselines with no prompt augmentation in Figure~\ref{fig:Ablation_ant_dit_ood}. The large variance of Prompt-DT may come from the large variance in episode returns across different testing tasks and the increased sparsity of training tasks.



\section{Conclusion}
\label{conclusion}

In this work, we proposed Prompt-based Decision Transformer to solve offline few-shot RL problems.
We empirically evaluated our algorithm and found it outperform the state-of-the-art offline meta-RL algorithm MACAW \cite{mitchell2021offline} in multiple benchmark domains.
We also showed that Prompt-DT is robust to the prompt length changes when trained with an expert dataest, while it is sensitive to the quality of the data provided in the prompt.

To the best of our knowledge, this is the first application of sequence-prediction models in the offline few-shot RL setting, developed based on Decision Transformer \cite{chen2021decision}.
Our algorithm is simple to implement, which only involves training a prompt-based Transformer, as opposed to training policy and value networks separately using an actor-critic algorithm in MACAW \cite{mitchell2021offline}.

We hope this work will inspire more investigation of applications of sequence-prediction models in RL.
In future work,
we consider designing objective functions that balance the weight of the trajectory prompt and the history context as we currently deem the length of the prompt as a hyperparameter,
and use prompt-based Transformer for other RL tasks like meta-imitation learning \cite{duan2017one,finn2017one}. We also notice that when using prompts subsampled from expert trajectories, Prompt-DT and Prompt-MT-BC fail to generalize in Meta-World's ML10 benchmark.
This motivates future works to design better prompts and prompt-based algorithms to solve complex compositional tasks.

\paragraph{Acknowledgements.} This work was supported by MIT-IBM Watson AI Lab and its member company Nexplore, and DARPA Machine Common Sense program. 
The information, data, or work presented herein was also funded by the Advanced Research Projects Agency-Energy (ARPA-E), U.S. Department of Energy, under Award Number DE-AR0001210. 
MDX and ZD gratefully acknowledge support from the National Science Foundation under grant CAREER CNS-2047454. The views and opinions of authors expressed herein do not necessarily state or reflect those of the United States Government or any agency thereof.

\newpage

\bibliography{citation}
\bibliographystyle{icml2022}


\newpage
\appendix
\onecolumn


\icmltitle{Appendix: Prompting Decision Transformer for Few-Shot Policy Generalization}

We provide the hyperparameters for Prompt-DT and baselines in Section~\ref{sec:appendix_hyp}, more experiment details in Section~\ref{sex:exp_detail} and ablation studies in Section~\ref{sec:appendix_ablation}.

\section{Hyperparameters}
\label{sec:appendix_hyp}
We show the hyperparameter of Prompt-DT and its variants in Table~\ref{tab:Hyperparameters_prompt} and Table~\ref{tab:Hyperparameters_prompt_env}, and MACAW in Table~\ref{tab:Hyperparameters_macaw}.
\begin{table}[h]
    \caption{Common Hyperparameters of Prompt-DT, Prompt-MT-BC, MT-ORL and MT-BC-Finetune}
    \label{tab:Hyperparameters_prompt}
    \centering
    \begin{tabular}{ll}
      \toprule
      Hyperparameters & Value \\
      \midrule
      $K$ (length of context $\tau$)  & 20 \\
      training batch size for each task   & 8 \\
      number of evaluation episodes for each task  & 20 \\
      learning rate & 1e-4 \\
      learning rate decay weight & 1e-4\\
      number of layers & 3 \\
      number of attention heads & 1 \\
      embedding dimension & 128 \\
      activation & ReLU \\
      \bottomrule
    \end{tabular}
    \vspace{-0.2in}
\end{table}

\begin{table}[h]
    \caption{Environment-specific Hyperparameters of Prompt-DT and Prompt-MT-BC}
    \label{tab:Hyperparameters_prompt_env}
    \centering
    \begin{tabular}{lcc}
      \toprule
      Environments & Target RewardS $G^{\star}$ & Prompt Length $K^{\star}$ \\
      \midrule
      Cheetah-dir & 1000 & 5 \\
      Cheetah-vel & 0 & 5  \\
      Ant-dir & 500 & 5  \\ 
      Dial & 10 & 15  \\
      Meta-World reach-v2 & 1500 & 2  \\
      \bottomrule
    \end{tabular}
    \vspace{-0.2in}
\end{table}

\begin{table}[h]
    \caption{Hyperparameters of MACAW}
    \label{tab:Hyperparameters_macaw}
    \centering
    \begin{tabular}{ll}
      \toprule
      Hyperparameters & Value \\
      \midrule
      training inner batch size & 256 \\
      evaluation batch size & prompt length $K^{\star}$ \\
      inner policy learning rate & 1e-3 \\
      inner value learning rate & 1e-3 \\
      outer policy learning rate & 1e-4 \\
      outer value learning rate & 1e-5 \\
      replay buffer size & 20k \\
      inner buffer size & 20k \\
      MLP net width & 300\\
      MLP net depth & 3\\
      activation & tanh \\
      adaptation step & 10 \\
      \bottomrule
    \end{tabular}
\end{table}

\newpage

\section{Experiment Details}
\label{sex:exp_detail}

We show the task index of the training and testing set for when evaluating the in-distribution generalization capability in Table~\ref{tab:set_meta}. In other words, the experiments in Section~\ref{sec:exp_prompt_main}, Section~\ref{sec:ablation_prompt_length} and Section~\ref{sec:ablation_prompt_quality} follows the training and testing division in Table~\ref{tab:set_meta}.

\begin{table}[h]
    \caption{Training and testing task indexes when testing the generalization ability in in-distribution tasks }
    \label{tab:set_meta}
    \centering
    \begin{tabular}{ll}
      \toprule
      \multicolumn{2}{c}{Cheetah-dir} \\
      \midrule
      Training set of size 2 & [0,1] \\
      Testing set of size 2 & [0.1]\\
      \midrule
      \multicolumn{2}{c}{Cheetah-vel} \\
      \midrule
      Training set of size 35 & [0-1,3-6,8-14,16-22,24-25,27-39]\\
      Testing set of size 5 & [2,7,15,23,26]\\
      \midrule
      \multicolumn{2}{c}{ant-dir} \\
      \midrule
      Training set of size 45 &  [0-5,7-16,18-22,24-29,31-40,42-49]\\
      Testing set of size 5 &  [6,17,23,30,41]\\
      \midrule
      \multicolumn{2}{c}{Meta-World reach-v2} \\
      \midrule
      Training set of size 15 &  [1-5,7,8,10-14,17-19]\\
      Testing set of size 5 &  [6,9,15,16,20]\\
      \midrule
      \multicolumn{2}{c}{Dial} \\
      \midrule
      Training set of size 6 &  Target pin number: [1,2,3,4,5,8]\\
      Testing set of size 4 &  Target pin number: [6,7,9,0]\\
      \bottomrule
    \end{tabular}
\end{table}

We also show and the task indexes when evaluating the out-of-distribution generalization capability in Table~\ref{tab:set_meta_ood} which accounts for the experiments in Section~\ref{sec:ablation_ood}.

\begin{table}[h]
    \caption{Training and testing task indexes when testing the generalization ability in out-of-distribution tasks}
    \label{tab:set_meta_ood}
    \centering
    \begin{tabular}{ll}
      \toprule
      \multicolumn{2}{c}{ant-dir} \\
      \midrule
      Training set of size 8 &  [8,13,16,20,22,26,32,37] \\
      Testing set of size 3 &  [1,4,41] \\
      \bottomrule
    \end{tabular}
\end{table}

\newpage

\section{Ablation Study}
\label{sec:appendix_ablation}

In this section, we provide addition ablation studies on the effect of prompt length in Section~\ref{sec:ablation_prompt_length_appendix}

\subsection{The Effect of Prompt Quantity}
\label{sec:ablation_prompt_length_appendix}
\begin{figure*}[h]
    \begin{center}
    \centerline{\includegraphics[width=0.95\linewidth]{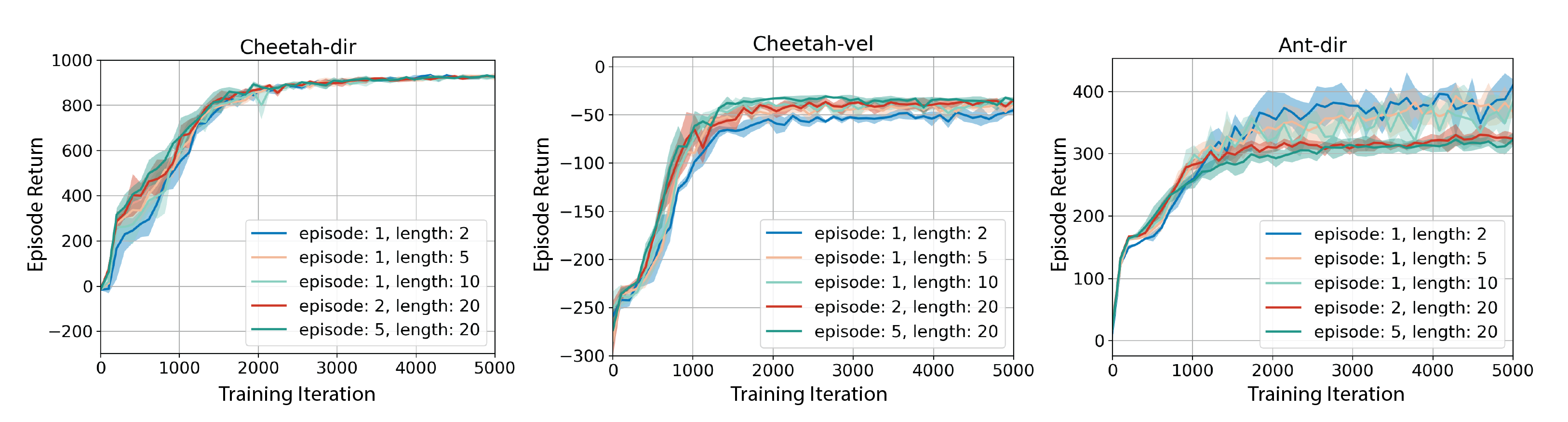}}
    \vskip -0.15in
    \caption{Ablation: The effect of trajectory prompt length to Prompt-DT's performance. Each plot is run with 3 seeds.}
    \label{fig:ablation_prompt_length_prompt_dt}
    \end{center}
    \vskip -0.2in
\end{figure*}

\begin{figure*}[h]
    \begin{center}
    \centerline{\includegraphics[width=0.95\linewidth]{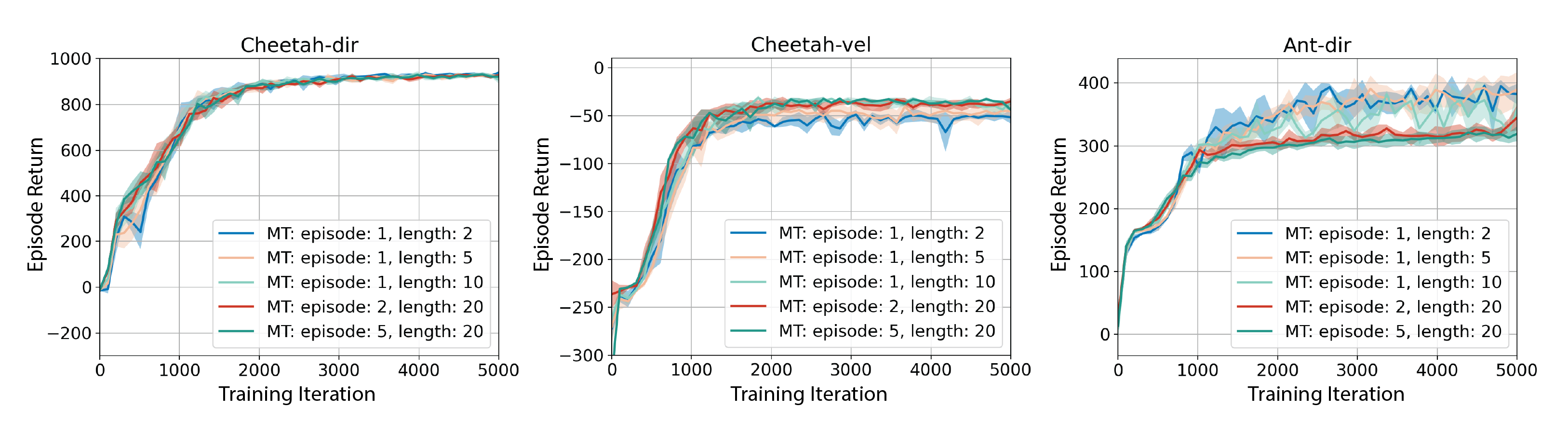}}
    \vskip -0.15in
    \centering
    \caption{Ablation: The effect of trajectory prompt length to Prompt-MT-BC's performance. Each plot is run with 3 seeds.}
    \label{fig:ablation_prompt_length_prompt_mt_bc}
    \end{center}
    \vskip -0.2in
\end{figure*}

We show evaluation curves along the training process with various prompt quantity. The training curves for Prompt-DT are shown in Figure~\ref{fig:ablation_prompt_length_prompt_dt} and curves for Prompt-MT-BC in Figure~\ref{fig:ablation_prompt_length_prompt_mt_bc}.
Both figures show that in Cheetah-dir, Cheetah-vel and Ant-dir, prompt-based methods are not sensitive to the number of episodes stored in the trajectory prompt or the prompt length.

\subsection{The Effect of Finetune Data's Quantity on MT-BC-Finetune}
\label{sec:ablation_finetune_data_MT_BC_Finetune}
\begin{figure*}[!htb]
    \begin{center}
    \centerline{\includegraphics[width=0.95\linewidth]{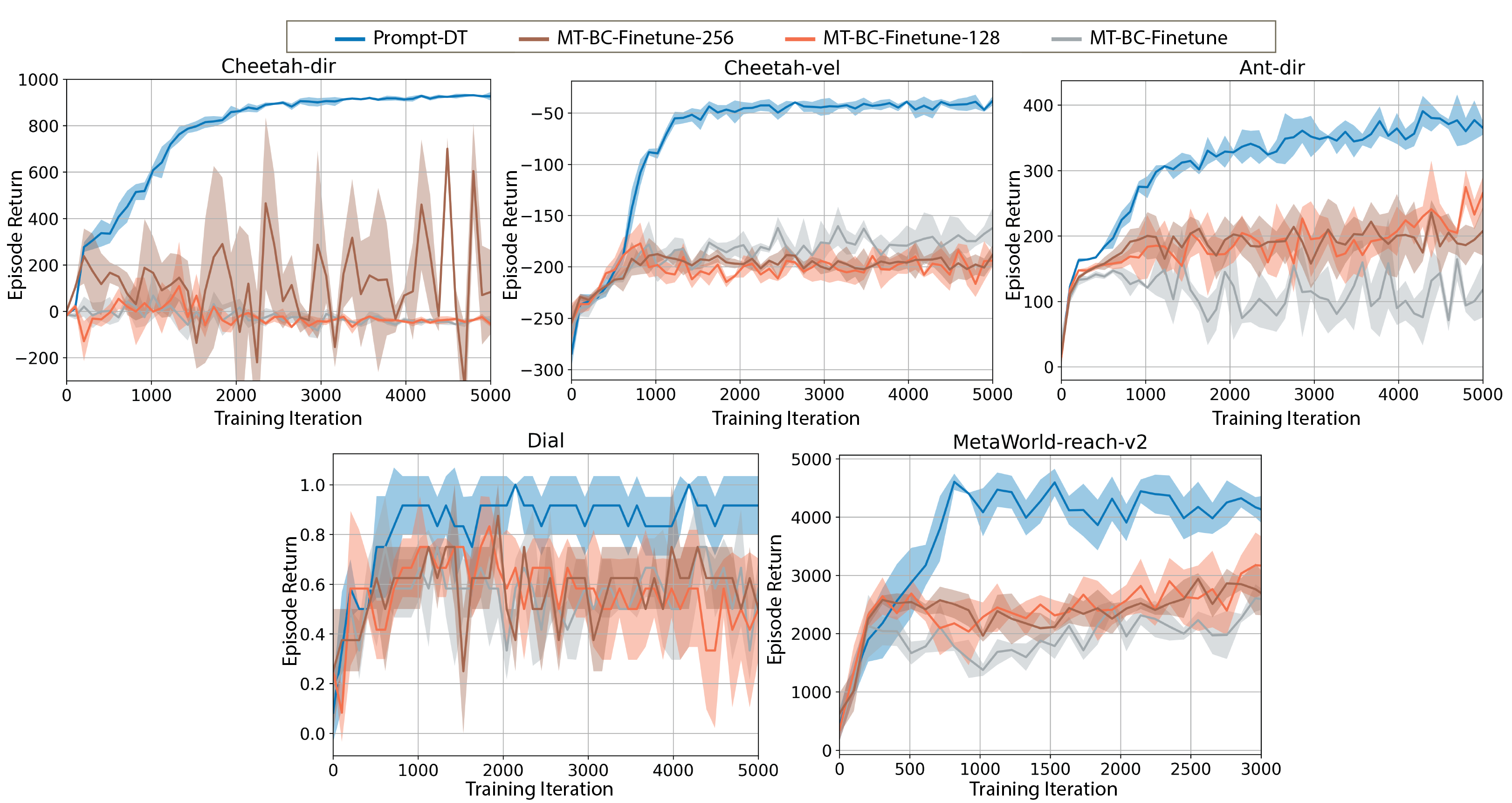}}
    \vspace{-0.15in}
    \caption{Ablation: The effect of finetune data's quantity on MT-BC-Finetune. MT-BC-Finetune-256 and MT-BC-Finetune-128 have a finetune batch size of 256 and 128 respectively.
    }
    \label{fig:ablation_finetune_data_MT-BC-Finetune}
    \end{center}
    \vspace{-0.2in}
\end{figure*}

We show MT-BC-Finetune's performance with various adaptation batch sizes in Figure~\ref{fig:ablation_finetune_data_MT-BC-Finetune}.
With limited finetune data, MT-BC-Finetune has difficulties in adapting to every task.
MT-BC-Finetune has a large performance variance in Cheetah-dir with an adaptation batch of size 256 and smaller variances in Cheetah-vel and Ant-dir. 
The large variance in Cheetah-dir may result from the disjoint state distribution in the two tasks with opposite rewards by design. 

Note that in Figure~\ref{fig:ablation_finetune_data_MT-BC-Finetune}, we use 10 adaptation steps. We notice that with 100 finetune steps and the adaptation batch size of 1280, MT-BC-Finetune could adapt to test tasks as shown in Table~\ref{tab:MT_BC_Finetune_oracle}.

\begin{figure*}[!htb]
    \begin{center}
    \centerline{\includegraphics[width=0.95\linewidth]{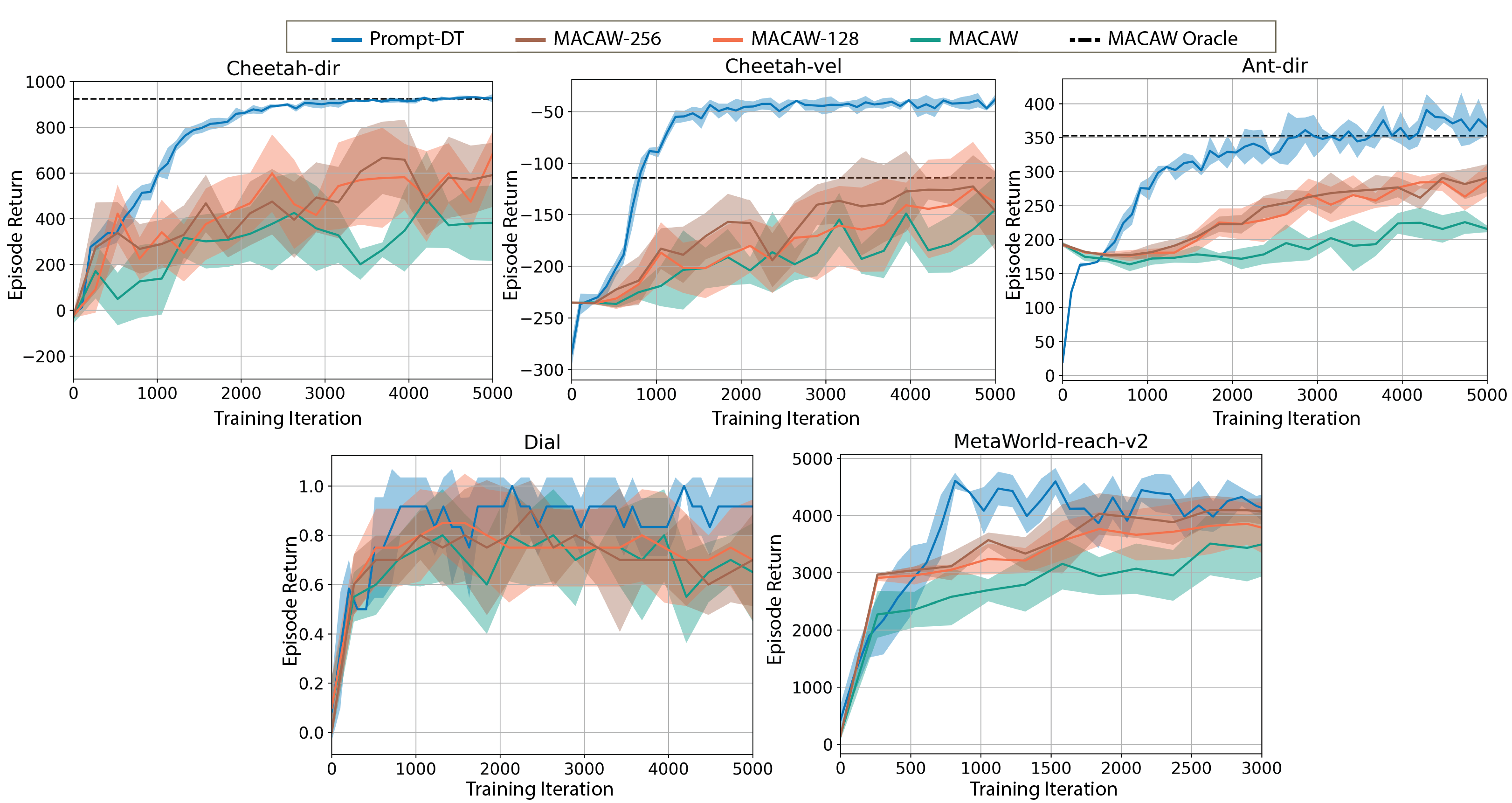}}
    \vspace{-0.15in}
    \caption{Ablation: The effect of finetune data's quantity on MACAW. MACAW-256 and MACAW-128 have a finetune batch size of 256 and 128 respectively.
    }
    \label{fig:ablation_macaw_data_MACAW}
    \end{center}
    \vspace{-0.3in}
\end{figure*}


\begin{table}[h]
    \centering
    \begin{tabular}{cccc}
    \toprule
        Environments & Adaptation Batch Size & Finetune Steps & Converged performance \\
        \midrule
        Cheetah-dir & 1280 & 100 & 930 \\
        Cheetah-vel & 1280 & 100 & -32 \\
        Ant-dir & 1280 & 100 & 470 \\
        \bottomrule
    \end{tabular}
    \caption{The performance of MT-BC-Finetune with adequate adaptation data and adaptation steps.}
    \label{tab:MT_BC_Finetune_oracle}
\end{table}

\subsection{The Effect of Finetune Data's Quantity on MACAW}
\label{sec:ablation_finetune_data_MACAW}

We show MACAW's performance with various adaptation batch sizes in Figure~\ref{fig:ablation_macaw_data_MACAW}. Each curve has 10 finetuning gradient steps. With an increasing adaptation batch size, MACAW's performance improves consistently across environments. However, MACAW-256 (MACAW with an adaptation batch size of 256) still underperforms Prompt-DT in Cheetah-dir, Cheetah-vel, and Ant-dir. In Meta-World reach-v2, Macaw-256 has a similar asymptotic performance with Prompt-DT but converges slower than Prompt-DT.




\newpage


\end{document}